\documentclass[letterpaper, 10 pt, conference]{ieeeconf}
\IEEEoverridecommandlockouts

\usepackage{times}

\usepackage[hyphens]{url}  %
\usepackage{graphicx} %
\usepackage{caption} %

\usepackage[style=ieee,backend=bibtex,url=false,doi=false,natbib=true,mincitenames=1,dashed=false,isbn=false]{biblatex}
\usepackage{multicol}
\usepackage[bookmarks=true]{hyperref}

\graphicspath{{../}{./}{./figures/}}
\usepackage[utf8]{inputenc}

\usepackage[font=small,labelfont=bf]{caption}

\usepackage{graphicx}
\usepackage{xcolor}
\usepackage[justification=centering]{subfig}
\usepackage{amsmath}
\usepackage{amsfonts}
\usepackage{amssymb}
\usepackage{mathtools}
\usepackage{todonotes}
\let\labelindent\relax
\usepackage{enumitem}
\usepackage{comment}
\usepackage{booktabs}
\usepackage[per-mode=symbol]{siunitx}
\DeclareSIUnit\minute{min}
\usepackage{tikzpagenodes}
\usepackage[noend]{algorithm,algorithmic}
\floatname{algorithm}{Algorithm}

\def\BibTeX{{\rm B\kern-.05em{\sc i\kern-.025em b}\kern-.08em
    T\kern-.1667em\lower.7ex\hbox{E}\kern-.125emX}}

\newcommand{\ph}[1]{{\textbf{#1:}}} %
\newcommand{\ncomment}[1]{}
\newcommand{\qcomm}[1]{{\color{red}Quest.}} %

\newcommand{\maximize}{\mathop{\mathrm{maximize}}}

\newcommand{\algname}{\text{FIG-OP}}

\newcommand{\globalirm}{\text{Topological Map}}
\newcommand{\globalirml}{\text{topological map}}

\newcommand{\riskmapl}{\text{local metric map}}

\newcommand{\graph}{G}
\newcommand{\node}{n}
\newcommand{\nodeset}{N}
\newcommand{\edge}{e}
\newcommand{\edgeset}{E}

\newcommand{\getedgeset}{\mathcal{E}}
\newcommand{\actioncost}{a}

\newcommand{\IGfunction}{\mathit{IG}}

\renewcommand{\path}{p}

\newcommand{\figfunction}{\mathit{FIG}}
\newcommand{\ffunction}{\mathit{F}}

\newcommand{\Gcost}{\rho^g}
\newcommand{\localmap}{M}
\newcommand{\Lcost}{\rho^\ell}
\newcommand{\pose}{m}

\usepackage{flushend}

\DeclareUnicodeCharacter{0301}{*************************************}

\begin{document}

\title{FIG-OP: Exploring Large-Scale Unknown \\ Environments on a Fixed Time Budget\\
\thanks{
The work is partially supported by the Jet Propulsion Laboratory, California Institute of Technology, under a contract with the National Aeronautics and Space Administration (80NM0018D0004), and Defense Advanced Research Projects Agency (DARPA), and funded through JPL’s Strategic University Research Partnerships (SURP) program.}
}

\author{
	Oriana Peltzer$^*$\thanks{*These authors contributed equally to this work.}\textsuperscript{\rm 1},
	\thanks{\textsuperscript{\rm 1}Department of Mechanical Engineering, Stanford University (e-mail: peltzer@stanford.edu).}  
    Amanda Bouman$^*$\textsuperscript{\rm 2}\thanks{\textsuperscript{\rm 2}Department of Mechanical and Civil Engineering, California Institute of Technology (e-mail: \{abouman, jwb@robotics\}@caltech.edu).},
    Sung-Kyun Kim\textsuperscript{\rm 3}\thanks{\textsuperscript{\rm 3}NASA Jet Propulsion Laboratory, California Institute of Technology (e-mail: \{sung.kim, aliahga\}\!@jpl.nasa.gov).}, \\
    Ransalu Senanayake\textsuperscript{\rm 4},
	\thanks{\textsuperscript{\rm 4}Department of Computer Science, Stanford University (e-mail: ransalu@stanford.edu).}
    Joshua Ott\textsuperscript{\rm 5}
    \thanks{\textsuperscript{\rm 5}Department of Aeronautics and Astronautics, Stanford University (e-mail: \{joshuaott, hdelecki, mykel, schwager\}\!@stanford.edu).},
    Harrison Delecki\textsuperscript{\rm 5},
    Mamoru Sobue\thanks{\textsuperscript{\rm 6}Graduate School of Frontier Sciences, University of Tokyo (e-mail: sobue.mamoru18\!@ae.k.u-tokyo.ac.jp). \newline \hspace*{1.1em} \copyright2022. All rights reserved.}\textsuperscript{\rm 6}, \\
    Mykel J. Kochenderfer\textsuperscript{\rm 5},
    Mac Schwager\textsuperscript{\rm 5},
    Joel Burdick\textsuperscript{\rm 2},
    Ali-akbar Agha-mohammadi\textsuperscript{\rm 3} \\
}

\maketitle

\begin{abstract}
We present a method for autonomous exploration of large-scale unknown environments under mission time constraints. We start by proposing the \textit{Frontloaded Information Gain Orienteering Problem} (\algname{}) -- a generalization of the traditional orienteering problem where the assumption of a reliable environmental model no longer holds. 
The \algname{} addresses model uncertainty by \emph{frontloading} expected information gain through the addition of a greedy incentive, effectively expediting the moment in which new area is uncovered. 
In order to reason across multi-kilometer environments, we solve \algname{} over an information-efficient world representation, constructed through the aggregation of information from a topological and metric map. Our method was extensively tested and field-hardened across various complex environments, ranging from subway systems to mines. In comparative simulations, we observe that the \algname{} solution exhibits improved coverage efficiency over solutions generated by greedy and traditional orienteering-based approaches (i.e. severe and minimal model uncertainty assumptions, respectively).

\end{abstract}

\IEEEpeerreviewmaketitle

\section{Introduction}\label{sec:intro}
Consider a time-limited mission where a robot, equipped with mapping and localization capabilities, is tasked with autonomously exploring a large-scale unknown environment. The robot must \textit{(i)} maintain an internal environment representation that encodes traversability and task-specific information, and \textit{(ii)} plan risk-mitigating paths that increase the robot's understanding of the world. All the while, the robot must account for motion and sensing uncertainty in order to plan and execute robust exploratory behaviors. 
To this end, we introduce a planning framework based on the \emph{orienteering problem} formulation \cite{orienteering} for solving the resource-constrained exploration problem over long horizons.

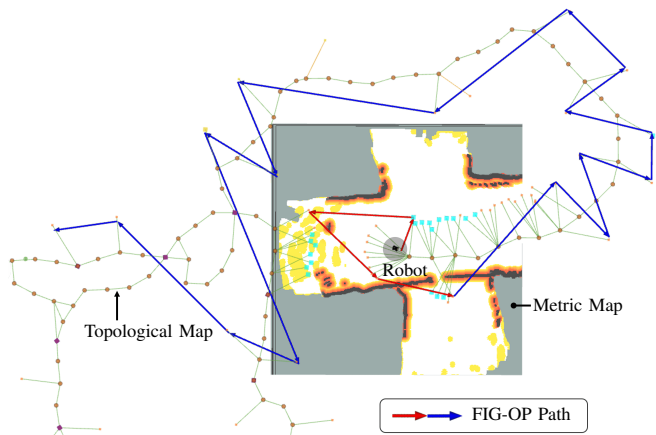
\begin{figure}[t!]
\centering
    \begin{tikzpicture}
    \node[anchor=south west,inner sep=0] (image) at (0,0) {\includegraphics[width=1.0\columnwidth]{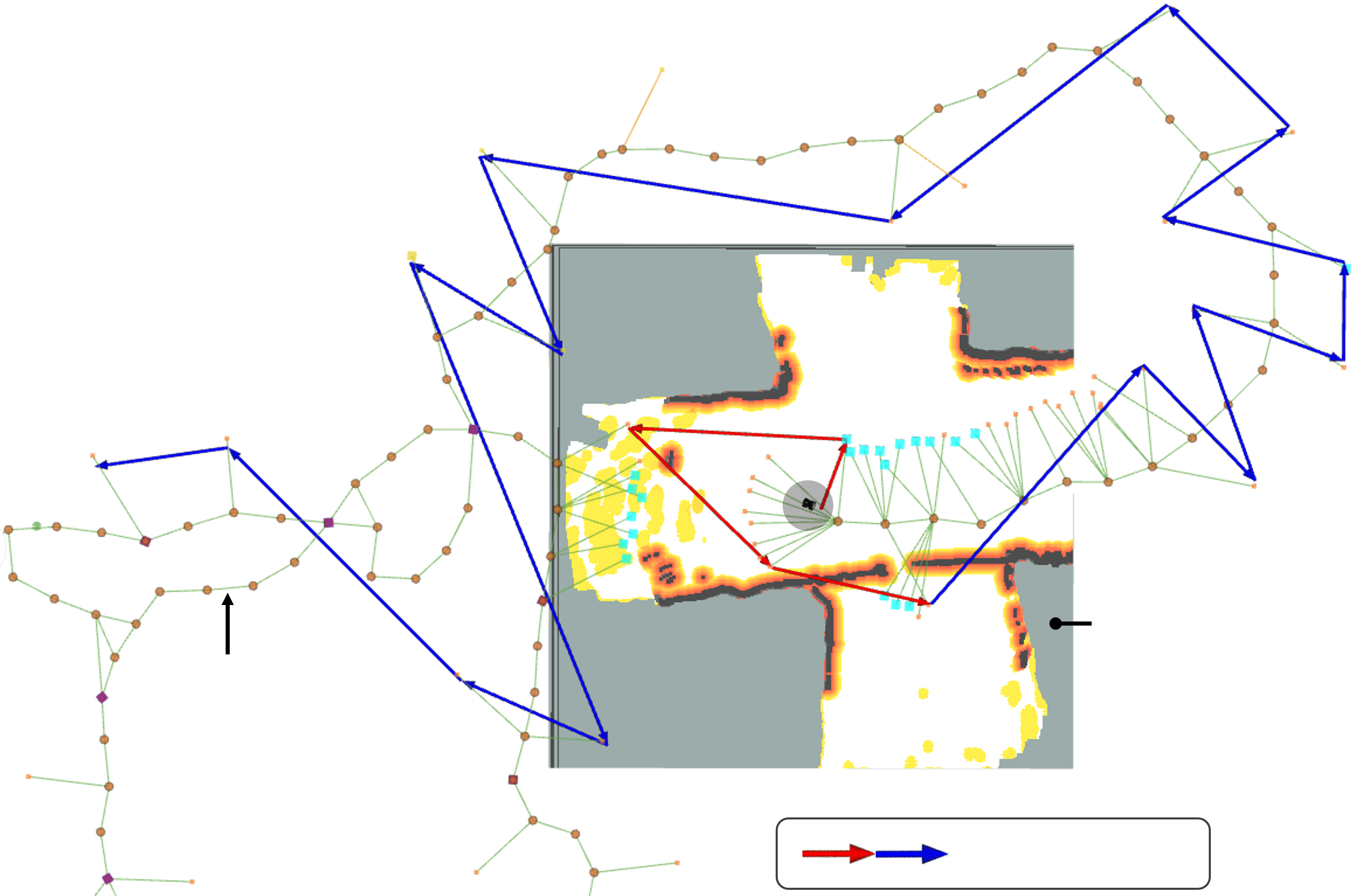}};
	    \begin{scope}[x={(image.south east)},y={(image.north west)}]
	    	\node [font=\scriptsize,above left,align=right,black] at (0.33,0.2) {\globalirm}; %
	    	\node [font=\scriptsize,above left,align=right,black] at (0.96,0.26) {Metric Map};
	    	\node [font=\scriptsize,above left,align=right,black] at (0.66,0.35) {Robot};
	    	\node [font=\scriptsize,above left,align=right,black] at (0.88,0.015) {FIG-OP Path};

	    \end{scope}
	\end{tikzpicture}	
  \caption{\algname{} path generated during the autonomous exploration of the Limestone Mines of the Kentucky Underground, Nicholasville, KY, with a Husky robot platform. The environment representation is broken down into two levels: a dense robot-centered \riskmapl{}, and a sparse globally-spanning \globalirml. Arrows show the FIG-OP solution. Path costs are computed on either the metric (red arrow) or topological map (blue arrow).} 
  \label{fig:main}
\end{figure}

Due to the computational constraints imposed by real-time systems, large-scale environments are commonly represented by topological graph-based structures.
Equipped with this representation, the majority of exploration-driven algorithms apply one-step lookahead strategies to extend the boundary of explored space, ultimately resulting in \emph{globally} sub-optimal plans. 
More recent approaches have found long horizon solutions using a traveling salesman problem formulation \cite{tare}. However, they assume unlimited mission time and complete understanding of the environment during policy execution. 

In reality, a topological graph structure representing an environment grows and changes dynamically as the robot uncovers new regions. 
In addition, the robot faces a risk of failure when traversing edges of the graph that is not accurately captured in the graph's action cost. As a result, the traditional orienteering problem formulation can suffer from 
overly ambitious planning, which can lead to a delay in the moment where new area is uncovered in favor of maximizing an unreliable reward estimate over the mission horizon. 
We argue, somewhat counter-intuitively, that a \emph{greedy} or \emph{time discounted} incentive allows for effectively exploring dynamic graphs, to ensure we extend the explored boundary in the near term. 

To address the challenges associated with optimal long term planning in unknown environments, we propose a framework consisting of two primary components. 
First, we construct a multi-fidelity world representation that encodes information about the robot's traversal risk and past sensor coverage. Then, we find long-horizon exploratory paths, robust to representation uncertainty, within the allotted mission time. Our specific contributions are as follows. 
\begin{enumerate}
    \item We construct a multi-fidelity, in terms of both time and space, world representation by combining risk and coverage information from \textit{(i)} a large-scale, but outdated and sparse, topological graph-based structure, and \textit{(ii)} a local-scale, but continuously updated and high resolution, metric grid-based structure. 

    By extracting information from both sources, 
    we increase the coverage rate by more than 35\% on average in a 30~minute run, when compared to using only the topological graph.
    \item We propose a variant of the orienteering problem, called the \textit{Frontloaded Information Gain Orienteering Problem} (\algname{}), for planning paths. The \algname{} objective is a function of both information gain and travel distance, resulting in solutions that shift, or \emph{frontload}, information gain earlier in time. We introduce an algorithm for solving \algname{} in real time based on Guided Local Search \cite{gls}. 
    \item The proposed solution was extensively tested on physical robots in various real-world environments. It also served as the top-level planner for team CoSTAR's entry in the Final Circuit of the DARPA SubT Challenge \cite{AliNeBula21}. In addition, we ran comparative experiments in high-fidelity simulation environments that show an improvement in coverage rate over competitive baseline methods.
\end{enumerate}

\section{Related Work}\label{sec:relatedwork}
The objective of the exploration problem is to maximize sensor \textit{coverage} of an unknown environment for a given mission time. The term coverage designates the area swept out by a robot's sensor footprint \cite{choset2001coverage}.

Exploration schemes relying on the identification of the boundary between uncovered and covered space, regions termed \emph{frontiers}, was first proposed by Yamauchi \cite{yamauchi1997frontier}. Since then, frontiers have been used extensively throughout the exploration literature \cite{sungpilgrim, howard2006experiments, burgard2000collaborative, umari2017autonomous}. 
Traditional frontier-based approaches construct one-step lookahead policies that maximize a utility function, accounting for expected gain of information and the cost of motion \cite{yamauchi1997frontier, gonzalez2002navigation, fang2019autonomous, wang2019autonomous, dang2019graph, rouvcek2021system, williams2020online}. 

Several frontier-based approaches have incorporated \emph{art gallery problem} schemes in order to find coverage-optimal paths towards an unexplored boundary.
Here, the objective is to find the minimum number of viewpoints that collectively maximizes coverage \cite{faigl2013determination, tare, heng2015efficient}. To ensure computational efficiency, Heng \cite{heng2015efficient} and Cao et al. \cite{tare} approximate the coverage problem, which is inherently submodular \cite{singh2009efficient, binney2010informative}, as a modular orienteering problem and traveling salesman problem, respectively. While non-myopic, these approaches are limited to a local region around the robot and assume no uncertainty in coverage information during policy construction.

\begin{figure}[t]

    \centering
    
    \subfloat[Orienteering Solution]{%
      \includegraphics[clip,width=0.48\columnwidth]{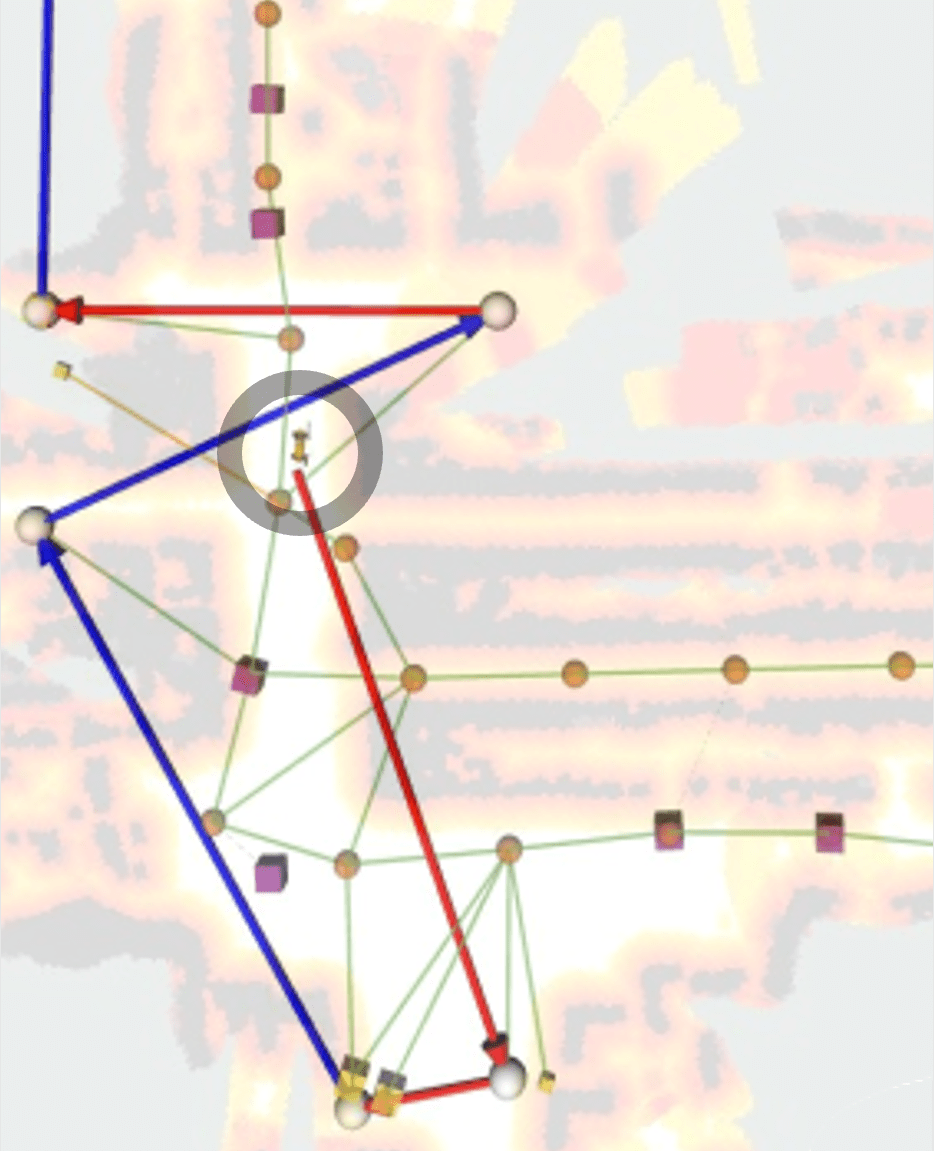}%
    } \,
    \subfloat[FIG Orienteering Solution]{%
      \includegraphics[clip,width=0.48\columnwidth]{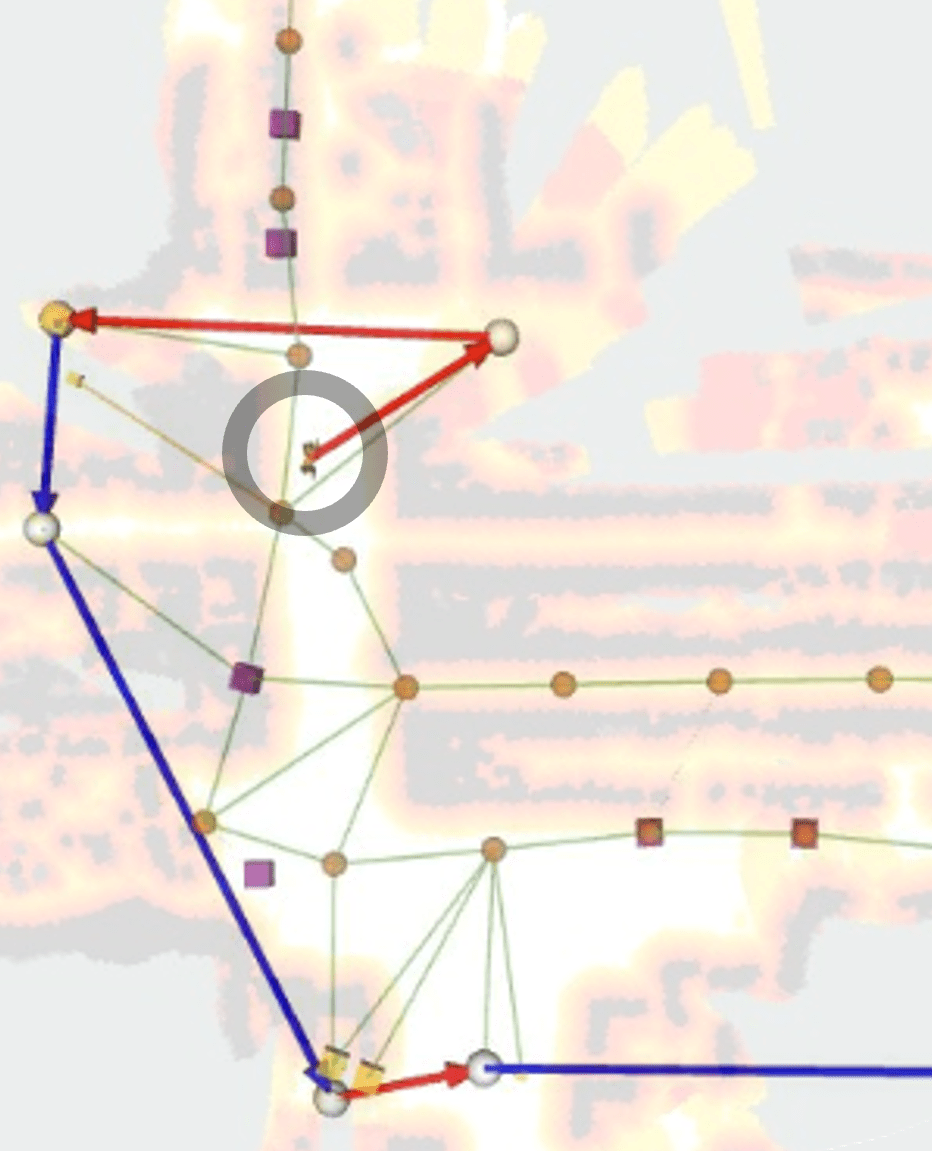}%
    }
    
    \caption{Comparison of the OP and $\algname$ solutions generated during Spot robot's autonomous exploration of a storage facility. Note that FIG-OP frontloads information gain, i.e. the first frontier in the solution path returned by $\algname$ has a lower action cost than that of the first frontier in the OP solution. By following the $\algname$ path, the robot will rapidly expose a new corridor in the library.}
    \label{fig:comparison}
\end{figure}

In order to address real-world stochasticity, the coverage problem has also been formulated as a Partially Observable Markov Decision Process (POMDP) \cite{Lauri2016planning, sungpilgrim}. POMDP solvers find robust long-horizon policies, but require a model of the robot's sequential action-observation process under motion and sensing uncertainty. While models have been effectively developed for short-range planning, POMDP planning at multi-kilometer scales suffer from model inaccuracy. 
As a result, \cite{sungpilgrim} assumes no changes to the state of the environment during global policy construction, significantly simplifying the problem to one in an MDP setting where frontiers are terminal states. As a consequence, the planning horizon is severely limited, resulting in a myopic exploration policy.

The orienteering problem provides long-horizon solutions to resource-constrained problems \cite{orienteering}.
Variants of the orienteering problem have been proposed for centralized multi-robot coverage problems \cite{liu2021team} and persistent monitoring problems \cite{yu2016correlated}.
However, these methods are not exploration-driven and, thus, rely on a prior map of the environment. When applied to the exploration problem, the OP suffers from many sources of model uncertainty (e.g. sensor measurements, hazard assessment, localization, and motion execution). At a high level, we focus on the false assumption that the graph structure is preserved over the planing horizon. On the contrary, during exploration, a robot will uncover new swaths of the environment, extending the horizon of explored space and augmenting its understanding of the world. 

Next, to address the aforementioned model uncertainty, we introduce an OP variant, FIG-OP, where the objective is designed to shift information gain earlier in time while simultaneously maintaining the long-term efficiency of the traditional OP.

\section{Problem Formulation}\label{sec:formulation}

Given an \emph{a priori} unknown environment, our objective is to construct a long-horizon path that provides global guidance to frontier regions so as to maximize information gathered over a predefined time budget. To solve this problem, we propose a variant of the Orienteering Problem
that produces solutions that \emph{frontload} information. 
We call this framework the \emph{Frontloaded Information Gain Orienteering Problem (\algname{})}.

We assume a graph-based environment representation $\graph=~(\nodeset,~\edgeset)$ 
with nodes $\nodeset$ and edges $\edgeset$. Nodes are discrete areas in space that represent frontiers, and edges represent actions. 
More precisely, we define an action to be a traversal from node $\node_i\in\nodeset$ to a node $\node_j\in\nodeset$, connected by an edge $\edge_{ij}\in\edgeset$.
Each edge $e$ in the graph has an action cost $\actioncost(\edge)$. 

Path $\path$ is a non-repeating sequence of nodes. $\mathcal{N}(\path)$ and $\getedgeset(\path)$ denote the set of nodes and edges along path $\path$, respectively. 
We define $a_\path(\node_i)$ to be the total action cost associated with traversal from the root node $\node_0$ to node $\node_i$ along the edges in $\path$.
More specifically, for any node $\node_i \in \path$, we define 
        $$a_\path(\node_i) = \sum_{\edge \in \getedgeset(\path_{0:i})} \actioncost(\edge),$$
where $a(e)$ is the action cost associated with edge $e$, and $\path_{0:i}$ is the contiguous subsequence of $\path$ from the root node $\node_0$ to node $\node_i$.

When visiting a node $\node_{i}$ on graph \graph, the robot collects information gain $\IGfunction(\node_{i}) \geq 0$, i.e., the amount of new area uncovered.
Let us now consider a \emph{frontloading} function $\ffunction$ that inflates information gain based on the accumulated action cost until the time of collection.
We wish to solve the optimization problem for \algname{} as:
\begin{equation}
\label{eq:frontloading}
\begin{aligned}
\maximize_{\path} &\quad \sum_{\node_i \in \mathcal{N}(\path)} \underbrace{{\boldsymbol{F} \big( 
\actioncost_\path(\node_i))} \cdot \boldsymbol{\IGfunction} \big(\node_i)}_{\figfunction}
\\[8pt]
\text{s.t.} &\quad \sum_{\edge \in \getedgeset(\path)}\actioncost(\edge) \leq \actioncost_{max}
  ~~ \text{and} ~~ \path_0 = \node_0,    \\
\end{aligned}
\end{equation}
\noindent where  $\actioncost_{max}$ is the robot's action cost budget.
The purpose of the function $\ffunction$ is to favor solutions where frontiers are visited in the near term. To this end, 
we define $\ffunction$ as follows:
\begin{equation}
\label{eq:FIG}
    F\big(a\big)= 1+k_1 \cdot S\bigg(\frac{a-k_2}{k_3}\bigg),
\end{equation}
\noindent where $S$ is the reversed logistic function $S(x)=\frac{1}{1+e^x}$, and amplitude $k_1$, inflection point $k_2$, and steepness $k_3$ are positive shaping parameters of $\ffunction$ (see Fig. \ref{fig:sensitivity})
The frontloading function $\ffunction$ exhibits the following properties that make it suitable for long-horizon exploration planning where model uncertainty is high.

\begin{figure}[t]
\centering
    \begin{tikzpicture}
	    \node[anchor=south west,inner sep=0] (image) at (0,0) {\includegraphics[width=0.9\columnwidth]{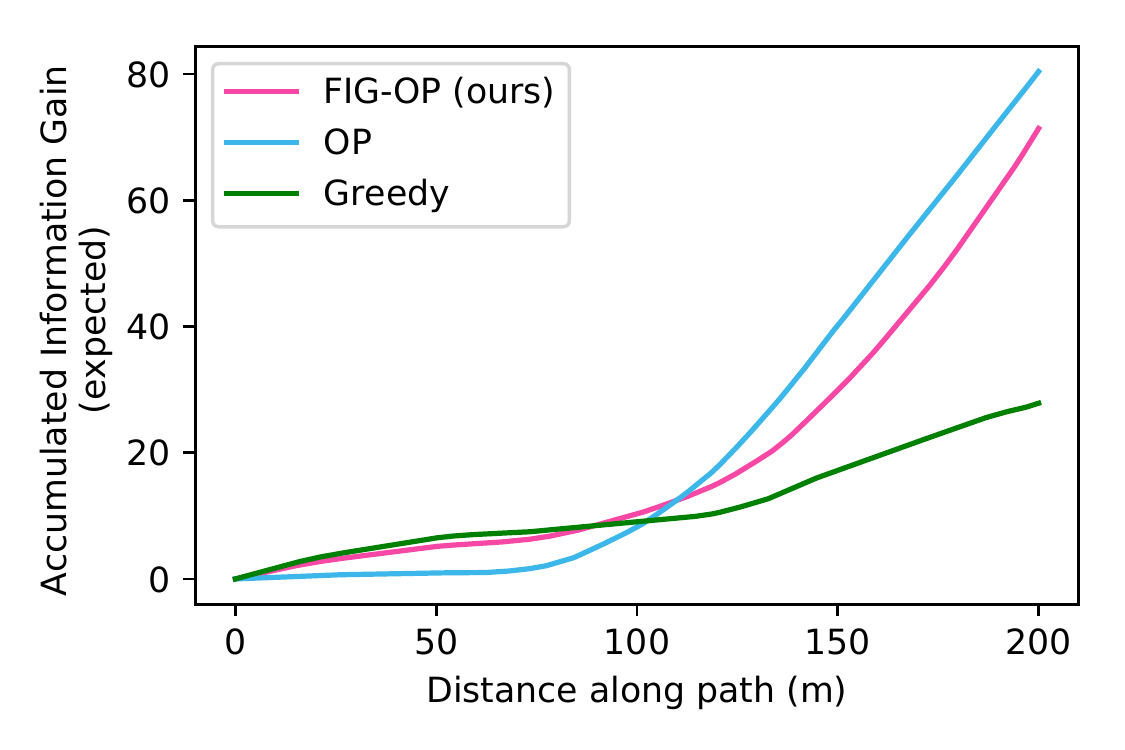}};
	\end{tikzpicture}	
\caption{Expected accumulated information gain for paths constructed according to FIG-OP, OP, and a greedy objective in the real-world mine (Fig. \ref{fig:hardware_husky_mine}). The curves are an average of 50 consecutive planning episodes with a planning horizon of 200~m. 
Early in the path, FIG-OP is competitive with the greedy algorithm, and thus quickly collects information gain. 
Later in the path, the greedy algorithm suffers due to an accumulation of globally suboptimal decisions. However, since FIG-OP encourages long-term efficiency, its solution stays competitive with the OP solution, which is designed to maximize information gain over the entire planning horizon.
}
\label{fig:expected_IG}
\end{figure}

\begin{enumerate}[label={\arabic*)}]
  \item \textit{Model Uncertainty Compensation}:
    To account for uncertainty in the time required to map the area beyond a frontier region,
    $\ffunction$ inflates information gain for frontiers within a local neighborhood of the robot,
    effectively expediting the moment in which new area is uncovered.
    As a result, policies are constructed according to a more optimistic model of the environment where frontiers likely lead to large unexplored swaths. By biasing towards short term information gain, we visit frontiers earlier in time with the expectation that they will significantly alter our world understanding by exposing new opportunities for information gathering. See Fig. \ref{fig:comparison}. 

  \item \textit{Long-Term Efficiency}:
    As action cost $\actioncost$ goes to infinity, $\ffunction$ converges to 1, reducing Eq.~(\ref{eq:frontloading}) to the standard OP formulation. By maintaining this long-term reward incentive, we assume a level of reliability in our environment model. We find that this policy-encoded foresight helps the robot gather more information over the mission time horizon. In essence, FIG-OP strikes a strategic balance between short and long-term information gain (Fig. \ref{fig:expected_IG}).

  \item \textit{Solution Regularization}:
    In contrast with the widely-adopted exponentially decaying function $E(\actioncost_\path)=\gamma^{\frac{\actioncost_\path}{k}}$,
    $\ffunction$ is most sensitive to action cost $a_\path$ at its inflection point $k_2$.
    This is a critical feature since local action costs, computed on the metric map are continuously updated, and thus are prone to fluctuations when estimates of traversability risk unexpectedly change (see Section \ref{sec:belief}). With this is mind, the inflection point $k_2$ can be selected to regularize the solution, i.e. lessen the path's susceptibility to locally fluctuating estimates. We find that this logistic form reduces detrimental oscillatory behavior and improves coverage performance (Fig.~\ref{fig:sensitivity}). 
  
\end{enumerate}

We solve the \algname{} using a receding horizon approach, where a model of the environment is constructed from the current state and used for planning at each iteration. By replanning regularly, the robot can adapt to unforeseen changes in the environment, such as newly uncovered areas or changing action costs.

\section{FIG-OP Representation}\label{sec:belief}

We introduce a multi-fidelity world representation for efficient evaluation of FIG-OP in Eq.~(\ref{eq:frontloading}). First, we define the two sources of information, namely \textit{Local Metric Maps} and \textit{Global Topological Maps} at the base of our methodology. Then, we introduce our proposed graph structure for effectively solving the \algname{} problem. After defining the information gain metric adopted, we provide our approach for computing action costs for the edges in the graph.

\subsection{Environment Representations}
The robot must always maintain an internal representation of its environment. Traditionally, these representations have fallen under one of two categories: \emph{topological} maps or \emph{metric} maps \cite{filliat2003map}. Topological maps are graph-based structures constructed by separating a space into a set of non-intersecting regions based on sensed features (e.g. frontiers). Since its resolution is dependent upon the complexity of the environment, topological maps are typically compact and scale well to large environments. Metric maps, meanwhile, are agnostic to environment complexity and divide the space into identical cells to form a grid structure.
When compared to topological maps, metric maps are easily constructed and maintained. However, since the grid must resolve detailed features in the environment, metric maps can suffer from large space and time complexities \cite{thrun1998learning}.
For compactness, versatility and fidelity, we extract and combine information from a local metric map and a global topological map for computation of the exploration reward and travel cost.

\begin{figure}[t]

    \centering
    
    \subfloat[][F objective shapes]{%
      \includegraphics[clip,width=0.5\columnwidth]{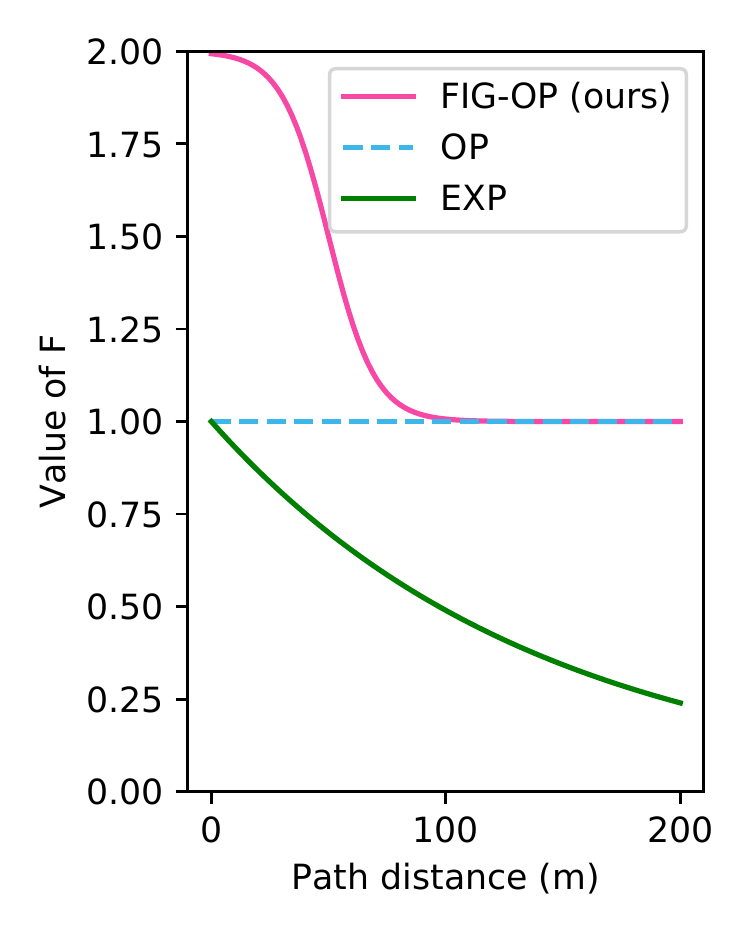}%
    }
    \subfloat[][Heading sensitivity comparison]{%
      \includegraphics[clip,width=0.53\columnwidth]{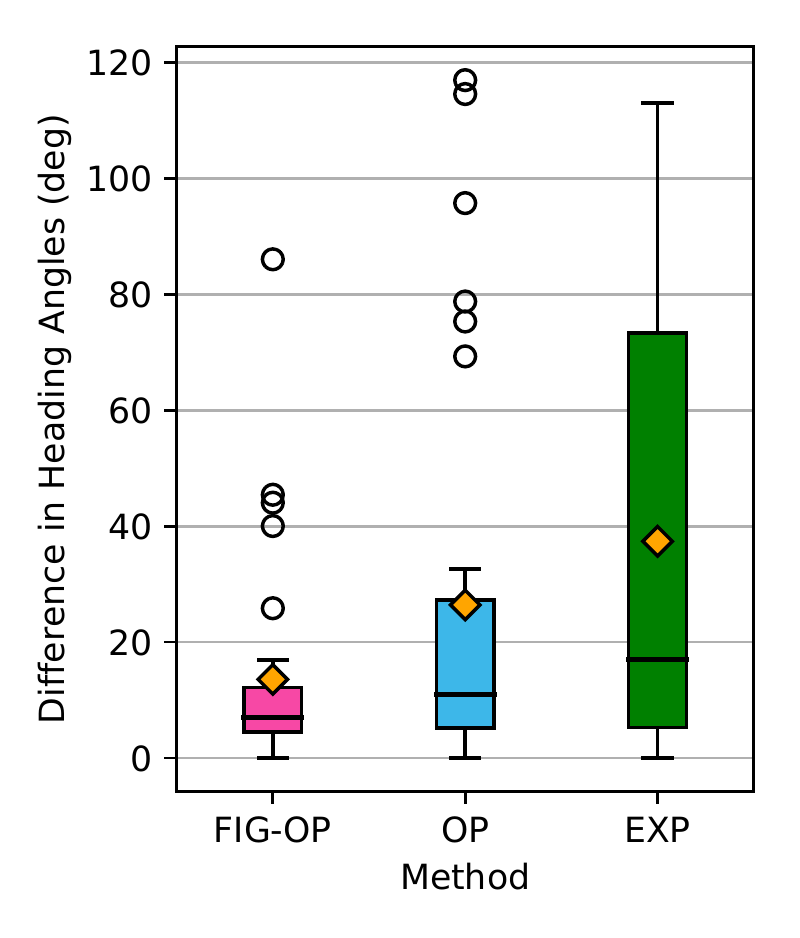}%
    } \,
    
    \caption{
    (a): The frontloading function $F$ in Eq.~(\ref{eq:FIG}) is shown for \algname{} (ours), orienteering problem (OP), and the exponentially discounted (EXP) problem objectives. \algname{} parameters are set to $k_1=1$, $k_2=50$, $k_3=10$. For EXP, we choose $k=50$ and $\gamma=0.7$, that empirically maximizes expected accumulated information gain over the 200 meter horizon. 
    (b): We compare the absolute difference in the robot's heading direction to the first waypoint in the path during consecutive planning episodes over a 30~s interval. This data is extracted from the Husky robot's exploration mission in a limestone mine where complex terrain causes unexpected changes in traversability risk estimates. 
    The plot illustrates the high sensitivity of the exponentially discounting objective (EXP) to locally changing risk estimates compared to our logistic objective form (FIG-OP). 
    Note: the box extends from the first quartile to the third quartile of the data, with a line at the median and a diamond at the mean.
    }
    \label{fig:sensitivity}
\end{figure}

\ph{Local Metric Map}
To capture local traversability risk at a high resolution, we employ a rolling, fixed-sized grid structure $\localmap=\{\pose_i\}$, which is centered at the robot's current position \cite{fan2021step}. This metric map, which is constructed by aggregating local point-cloud sensor measurements, captures mobility stressing features in the environment, such as obstacles, slopes, and ground roughness. We denote $\pose_i$
as a cell in the grid-based map. Then, using the geometric planner proposed by \citet{fan2021step}, we define $\Lcost(\pose_i, \pose_j)$ as the instantaneous cost of the risk-minimized path between cells $\pose_i$ and $\pose_j$.

\ph{Global Topological Map}
To capture the global exploration state at multi-kilometer scales, we use a sparse bidirectional graph structure $\graph^g=(\nodeset^g, \edgeset^g)$, formulated as the \emph{Global Information Roadmap} by \citet{sungpilgrim}. This topological map is globally fixed and consists of two mutually exclusive subsets of nodes: \emph{breadcrumbs} $n^g_{i,b} \in N^g$ and \emph{frontiers} $n^g_{i,f} \in N^g$. Breadcrumbs are generated in the robot's wake and capture covered traversable spaces in the environment. Alternatively, frontiers are generated at the boundary between explored and unexplored areas and, thus, capture uncovered traversable spaces. Given the topological map $G^g$, we define the action cost $\Gcost(\node^g_i, \node^g_j)$ of traversal between two nodes $\node^g_i$ and $\node^g_j$ as the distance associated with the shortest path, computed by applying Dijkstra's algorithm over the weighted graph. The edge weights are computed between two nodes within a neighborhood using the above geometric planner over the metric map. Due to range and computational limitations, edge weights are computed only once, despite changes in risk assessment over time.

\begin{figure}[!t]
\centering
    \begin{tikzpicture}
	    \node[anchor=south west,inner sep=0] (image) at (0,0) {\includegraphics[width=\columnwidth]{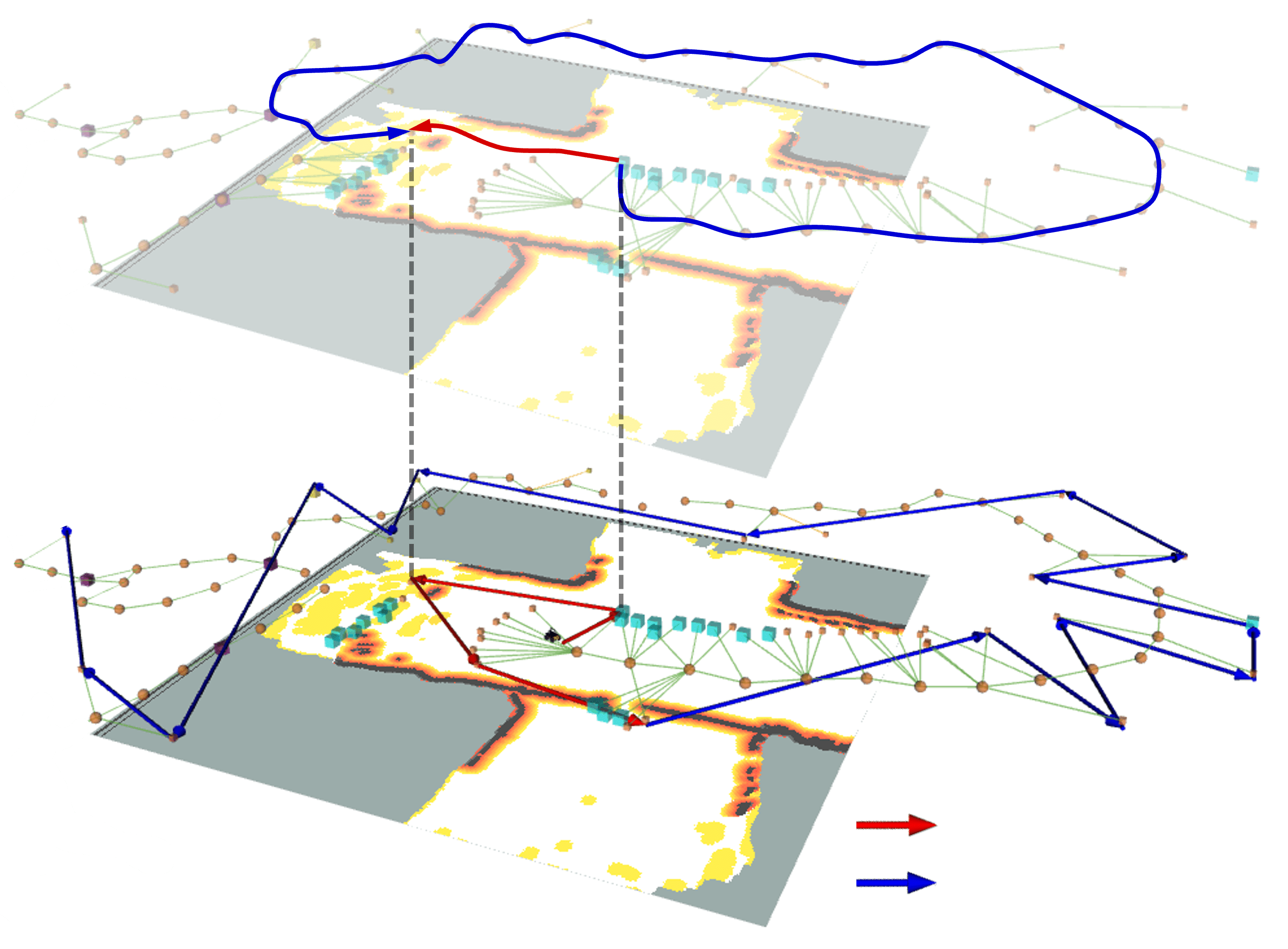}};
	    \begin{scope}[x={(image.south east)},y={(image.north west)}]
	    
        	\node [above left,align=right,black] at (0.865,0.10) {Metric};
        	\node [above left,align=right,black] at (0.95,0.025) {Topological};
        	
	    \end{scope}
	\end{tikzpicture}	
\caption{The FIG-OP graph combines low-fidelity action costs computed on the topological map (blue) and high-fidelity action costs computed on the metric map (red). Here, the topological-based and metric-based paths between two frontiers is displayed on the top layer. By using the more accurate metric-based path, FIG-OP finds a path that ``closes a loop" on the topological map -- a critical feature for accurately estimating action costs globally.} 
\label{fig:multifidelity}
\end{figure}

\subsection{FIG-OP Graph Structure}

We propose a multi-fidelity \emph{complete graph} structure $\graph = (\nodeset, \edgeset)$, which combines information from both the local metric map and the global topological map (Fig. \ref{fig:main}). To reduce the size of the search space, a node $\node_i \in N$ designates a cluster of frontiers in the topological map, determined using the DBSCAN algorithm \cite{ester1996density}. 
Functions $f_\ell()$ and $f_g()$ map a cluster centroid to a metric cell and topological node, respectively.
Every node pair $(\node_{i}, \node_{j})$ is connected by an edge $\edge_{ij} \in E$. 
Practically, we assume that by visiting one frontier in the cluster, any neighboring frontier in the cluster is accessible within a predefined distance. 

\subsection{Information Gain} \label{ssec:info_gain}
We define information gain $\IGfunction(\node^{g}_{i})$ to be the expected area that can be uncovered by reaching a frontier node $\node^{g}_i$ on the topological map.
Frontier segments are detected on the metric map using an omnidirectional LiDAR sensor at a range $r_{\text{sense}}$, based on the Fast Frontier Detector 
method \cite{keidar2012robot}.
Since the occupancy of cells beyond $r_{\text{sense}}$ cannot be reliability determined \cite{visser2007beyond}, we do not count the number of cells expected to be visible to the robot as it travels along an edge.
Instead, 
we define $\IGfunction(\node^{g}_i)$ as a function of the approximated breadth and depth of the uncovered area beyond a frontier node. Frontier breadth is estimated by measuring the length (i.e. number of metric map cells) of the frontier segment. Frontier depth is estimated by attempting to associate the frontier segment with segments detected at longer ranges using connected component analysis. Frontiers that can be associated with long-range segments have larger depth values.

\subsection{Multi-Fidelity Action Cost}

Since the topological map consists of sparsely sampled nodes and maintains edge weights based on outdated risk information, it can only provide a crude estimate of the traversal cost between two clusters $\node_i$ and $\node_j$. To overcome this weakness, we compute traversal costs over the metric map for clusters located within its bounds based on the most up-to-date traversal risk assessment.
We combine both metric and topological information to form a multi-fidelity cost estimate: 
\begin{equation}
\nonumber
    \begin{aligned}
    \actioncost(\edge_{ij}) = 
    \begin{cases}
    \Lcost\big(f_\ell(n_i), f_\ell(n_j) \big), \quad \text{ if } f_\ell(n_i), f_\ell(n_j) \in \localmap
    \\[7pt]
    \mathit{\Gcost}\big(f_g(n_i), f_g(n_j)\big), \quad \text{otherwise.}
    \end{cases} \\
    \end{aligned}
\end{equation}

\noindent Fig. \ref{fig:multifidelity} illustrates the benefits of combining action costs computed on both the metric and topological maps.

\section{\algname{} Solver}\label{sec:algorithm}

\begin{algorithm}[t]
\small
 \caption{FIG-GLS: Front-loading variant of Guided Local Search}
 \begin{algorithmic}[1] 
 \renewcommand{\algorithmicrequire}{\textbf{Input:}}
 \renewcommand\algorithmicthen{}
 \renewcommand\algorithmicdo{}
  \REQUIRE Graph $\graph$, previous solution $\mathcal{B}_{t-1}$, new frontiers $\mathcal{F}$
  \STATE $\mathcal{S} \gets $ Construct($\graph, \mathcal{B}_{t-1}, \mathcal{F}$) 
  \STATE $\mathcal{B}$.solution $\gets$ $\mathcal{S}$ 
  \STATE $\mathcal{B}$.cost $\gets$ cost($\mathcal{S}$) \; $\triangleright$ \textit{Using equation (\ref{eq:frontloading})}
  \STATE AlgLoop $\gets$ 0
  \WHILE{AlgLoop $\leq$ MaxAlg}
  \STATE AlgLoop $\gets$ AlgLoop + 1
  \STATE LsLoop $\gets$ 0
  
  \WHILE{Solution improved and LsLoop $\leq$ MaxLs}
  \STATE LsLoop $\gets$ LsLoop + 1
  \STATE $\mathcal{S} \gets$ TSP($\mathcal{S}$)
  \STATE \textbf{$\mathcal{S} \gets$ Swap($\mathcal{S}$)}
  \STATE \textbf{$\mathcal{S} \gets$ BackwardSwap($\mathcal{S}$)}
  \STATE $\mathcal{S} \gets$ Insert($\mathcal{S}$)
  \STATE $\mathcal{S} \gets$ Replace($\mathcal{S}$)
  \ENDWHILE
  
  \IF {cost($\mathcal{S}) > \mathcal{B}$.cost:}
  \STATE $\mathcal{B}$.solution $\gets$ $\mathcal{S}$
  \STATE $\mathcal{B}$.cost $\gets$ cost($\mathcal{S}$) \; $\triangleright$ \textit{Using equation (\ref{eq:frontloading})}

  \ELSIF {$\mathcal{S} = \mathcal{B}$}
  \IF {not disturbed before}
  \STATE $\mathcal{S} \gets$ Disturb($\mathcal{S}$)
  
  \ELSE
  \STATE \textbf{$\mathcal{S} \gets$ Swap($\mathcal{S}$)}
  \STATE \textbf{$\mathcal{S} \gets$ BackwardSwap($\mathcal{S}$)}
  \STATE \textbf{return } $\mathcal{B}$
  
  \ENDIF
  
  \IF{AlgLoop $=$ MaxAlg$/2$}
  \STATE $\mathcal{S} \gets$ Disturb($\mathcal{S}$)
  \ENDIF

  \ENDIF

  \ENDWHILE
  \STATE \textbf{return } $\mathcal{B}$
 \end{algorithmic} 
 \label{ftlalg}
 \end{algorithm}

Guided Local Search is a state-of-the-art heuristic method for solving the Orienteering Problem \cite{gls, orienteering}. We extend the method to allow for the modified OP objective function in Eq. (\ref{eq:frontloading}). Alg.~\ref{ftlalg} describes the search procedure at a high level. 
We introduce three notable changes to \cite{gls}:
\begin{enumerate}[label={\arabic*)}]
    \item Solutions are evaluated, i.e. the path cost is computed, according to the revised \algname{} objective function Eq.~(\ref{eq:frontloading}). See lines 3 and 17 in Alg.~\ref{ftlalg}.
    \item Two procedures, \textit{Swap} and \textit{BackwardSwap}, are introduced to explore different orderings of nodes within a solution and increase the objective value. \textit{Swap} iterates over the path in a forward direction. To encourage frontloading high information gain frontiers, \textit{BackwardSwap} considers switching nodes in reverse order. If the path objective remains the same, the swap is conducted if the total path cost is decreased. 
    \item Every iteration, we seed \algname{} with the previous solution updated with newly generated frontiers inserted at the front of the path.
\end{enumerate}

\noindent The complexity of this sub-optimal solver grows quadratically with the number of frontier clusters in the graph, which scales similarly to commonly adopted polynomial-time motion planners.

\section{Experiments}\label{sec:experiments}

\begin{table}[t]
\begin{center}
\caption{For each listed algorithm, the average coverage metric over 5 runs in a simulated maze and subway environment is displayed. Standard deviations are provided in parenthesis. 
}
\label{table:maze_results}
    \begin{tabular}{@{}lrrr@{}}
    \toprule
    \multicolumn{4}{ c }{\textbf{Simulated Maze}} \\
    \midrule
    Method & Coverage Rate & 30 min Coverage & Planning Time\\ 
    & (\si{\meter\squared\per\minute}) & (\si{\meter\squared}) & (\si{\second}) \\ \midrule
    \algname{} (ours) & 150.3 (7.4) & 4577 (165) & 0.22 (0.42)\\ 
    OP & 121.7 (6.3) & 3822 (275) & 0.12 (0.34) \\ 
    \textbf{Greedy} & \textbf{154.3 (8.0)} & \textbf{4646 (257)} & 0.02 (0.04) \\ 
    FIG-LF & 113.5 (7.0) & 3468 (250) & 0.15 (0.42) \\ 
    \midrule
    \multicolumn{4}{ c }{\textbf{Simulated Subway Station}} \\
    \midrule
    Method & Coverage Rate & 95\% Coverage Time & Planning Time\\ 
    & (\si{\meter\squared\per\minute}) & (\si{\minute}) & (\si{\second}) \\ \midrule
    \textbf{\algname{} (ours)} & \textbf{163.7 (20.5)} & \textbf{12.07 (1.53)} & 0.08 (0.08)\\ 
    OP & 142.8 (14.1) & 13.71 (0.98) & 0.05 (0.06) \\ 
    Greedy & 135.4 (17.2) & 14.64 (1.83) & 0.01 (0.01) \\ 
    FIG-LF & 130.8 (19.4) & 14.47 (2.15) & 0.09 (0.56) \\ 
    \bottomrule
    \end{tabular}
    \label{1}
\end{center}
\end{table}

We perform simulation studies and real-world experiments with a four-wheeled vehicle (Husky robot) and a quadraped (Boston Dynamics Spot robot) in order to evaluate our proposed algorithm. The robot is equipped with custom sensing and computing systems~\cite{Otsu2020,AliNeBula21arXiv,AutoSpot}, and the entire autonomy stack runs in real-time on an Intel Core i7 processor with 32 GB of RAM. The stack relies on a multi-sensor fusion framework, the core of which is 3D point cloud data provided by LiDAR range sensors~\cite{Ebadi2020}. To evaluate planner performance in both simulation and real-world, we compute \emph{coverage} [$m^2$] as the accumulated area within the robot's sensor footprint during a run. Throughout this section, we detail how our experimental findings validate core features of our proposed method.

\subsection{Simulations}

We demonstrate FIG-OP's performance in a simulated subway and maze environment.
We compare the performance of \algname{} against the following frontier-based exploration planners in the simulated subway and maze environments. 

\begin{enumerate}[label={\arabic*)}]
  \item \textit{\algname{}:}
    Proposed method where FIG-OP ($k_1=1$, $k_2=50$, and $k_3=10$) is solved over the multi-fidelity complete graph $G$ using Alg.~\ref{ftlalg}.
  \item \textit{FIG-OP with Low-Fidelity Action Costs (FIG-LF)}:
    Modification to the proposed method where all action costs are computed on the topological graph $G^g$.
  \item \textit{Greedy:}
	Myopic planner that selects the frontier with the smallest action cost based on $G$ \cite{yamauchi1997frontier}.
  \item \textit{OP}:
	Long-horizon planner where the orienteering problem is solved over $G$ using GLS algorithm in \cite{gls}.

\end{enumerate}

\noindent

\begin{figure}[t!]

    \centering
    
    \subfloat[][Simulated Subway]{\includegraphics[clip,width=0.8\columnwidth]{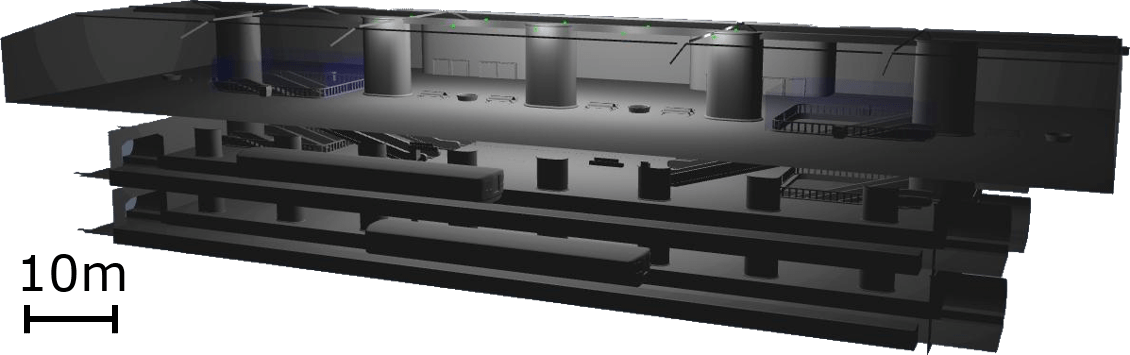}} \\
	
	\subfloat[][]{\includegraphics[clip,width=0.5\columnwidth]{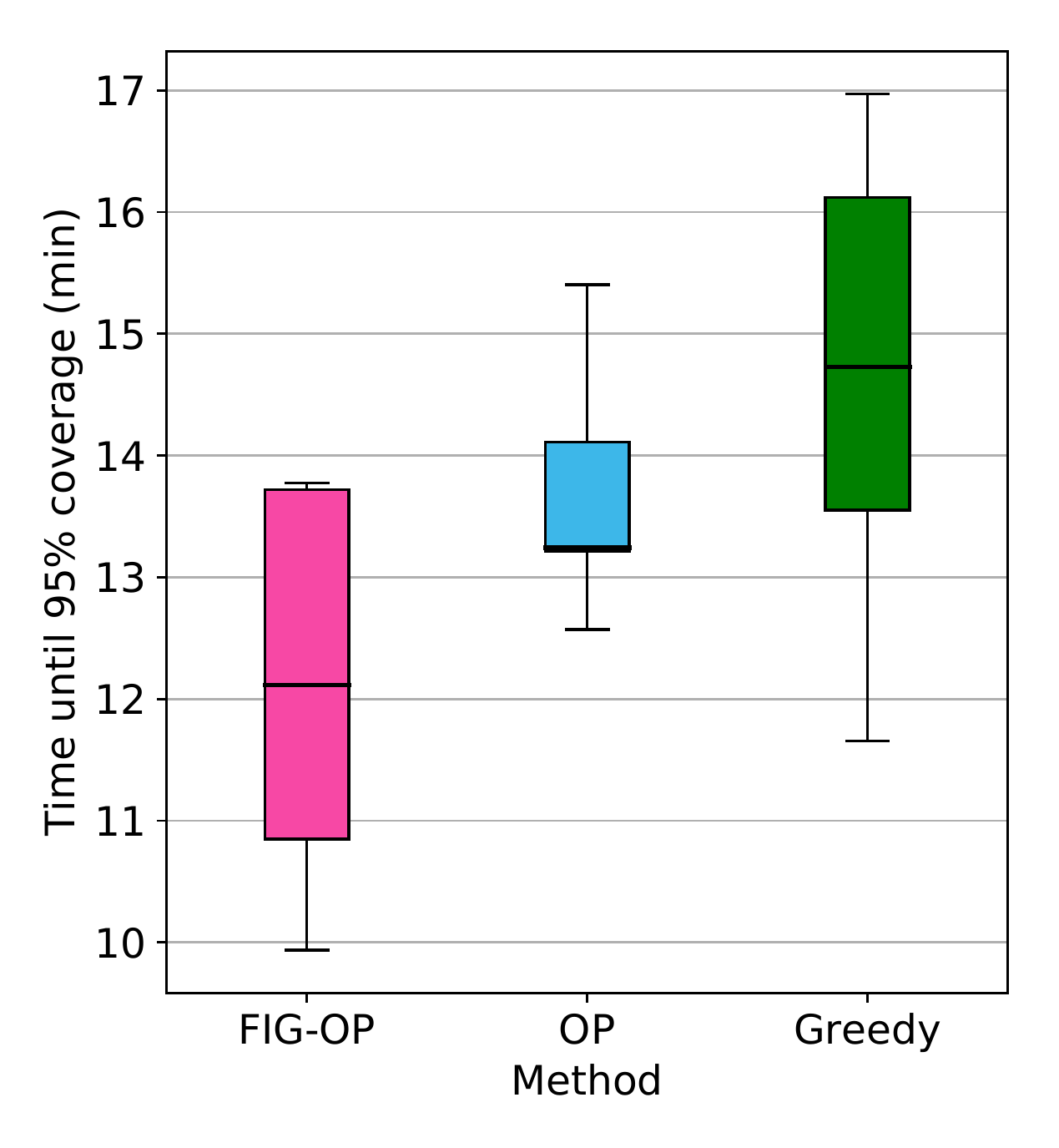}}
	\subfloat[][]{\includegraphics[clip,width=0.5\columnwidth]{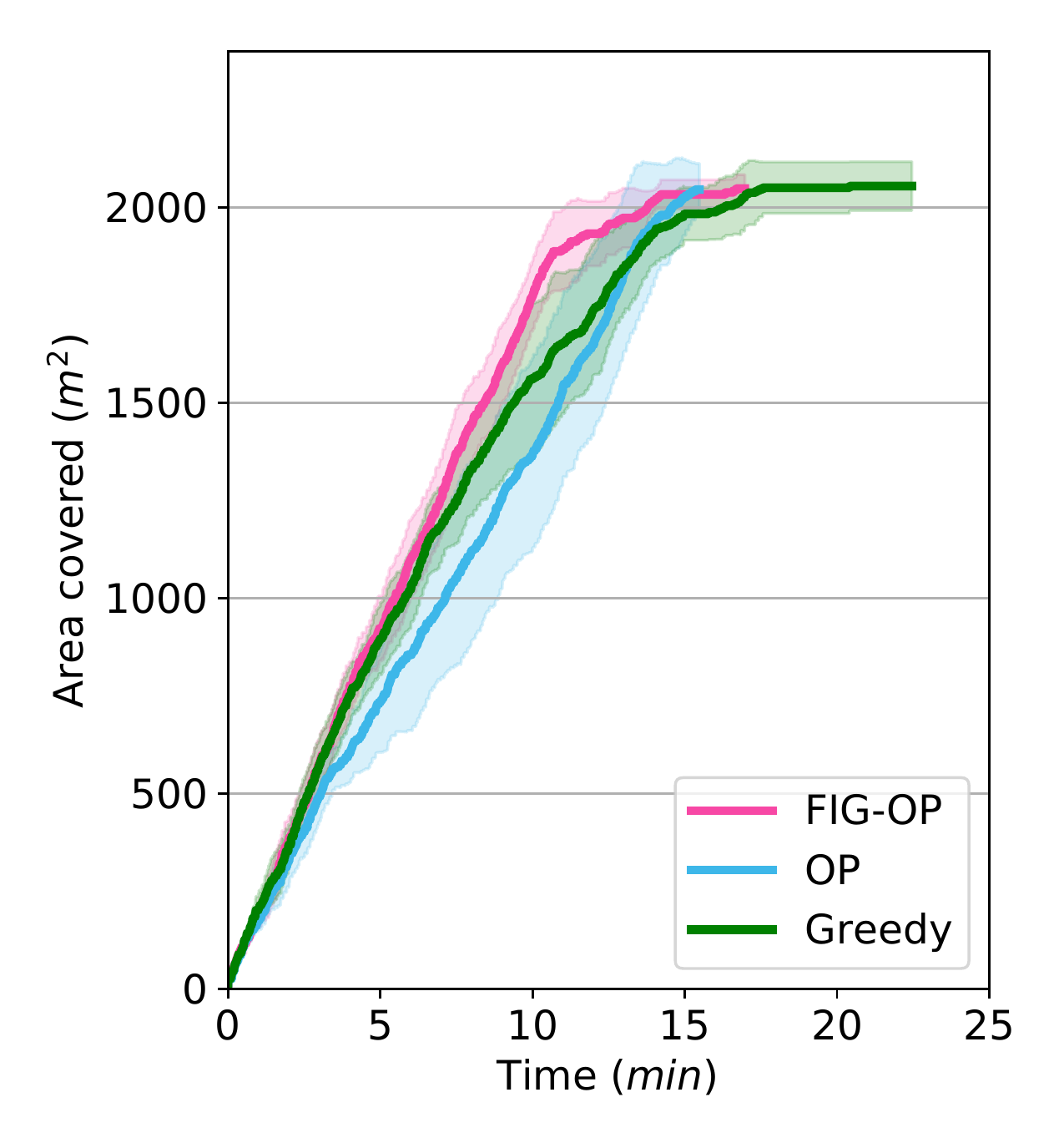}}	
    
	\caption{
	Exploration by our proposed \algname{} and baseline methods in a simulated subway. 
	The exploration metrics for each method consist of five runs, with curve (c) displaying the average. 
	Refer to Fig.~\ref{fig:sensitivity} for details about the box plot. }
    \label{fig:subway}
\end{figure}

\begin{figure}[t!]

    \centering
    
	\subfloat[][Simulated Maze]{\includegraphics[clip,width=0.55\columnwidth,  angle=90]{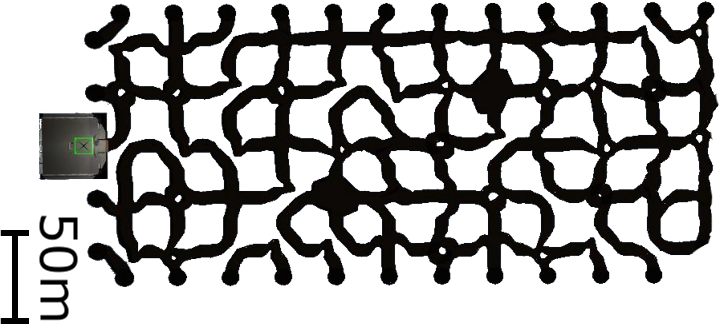}} \; \;
	\subfloat[][]{\includegraphics[clip,width=0.5\columnwidth]{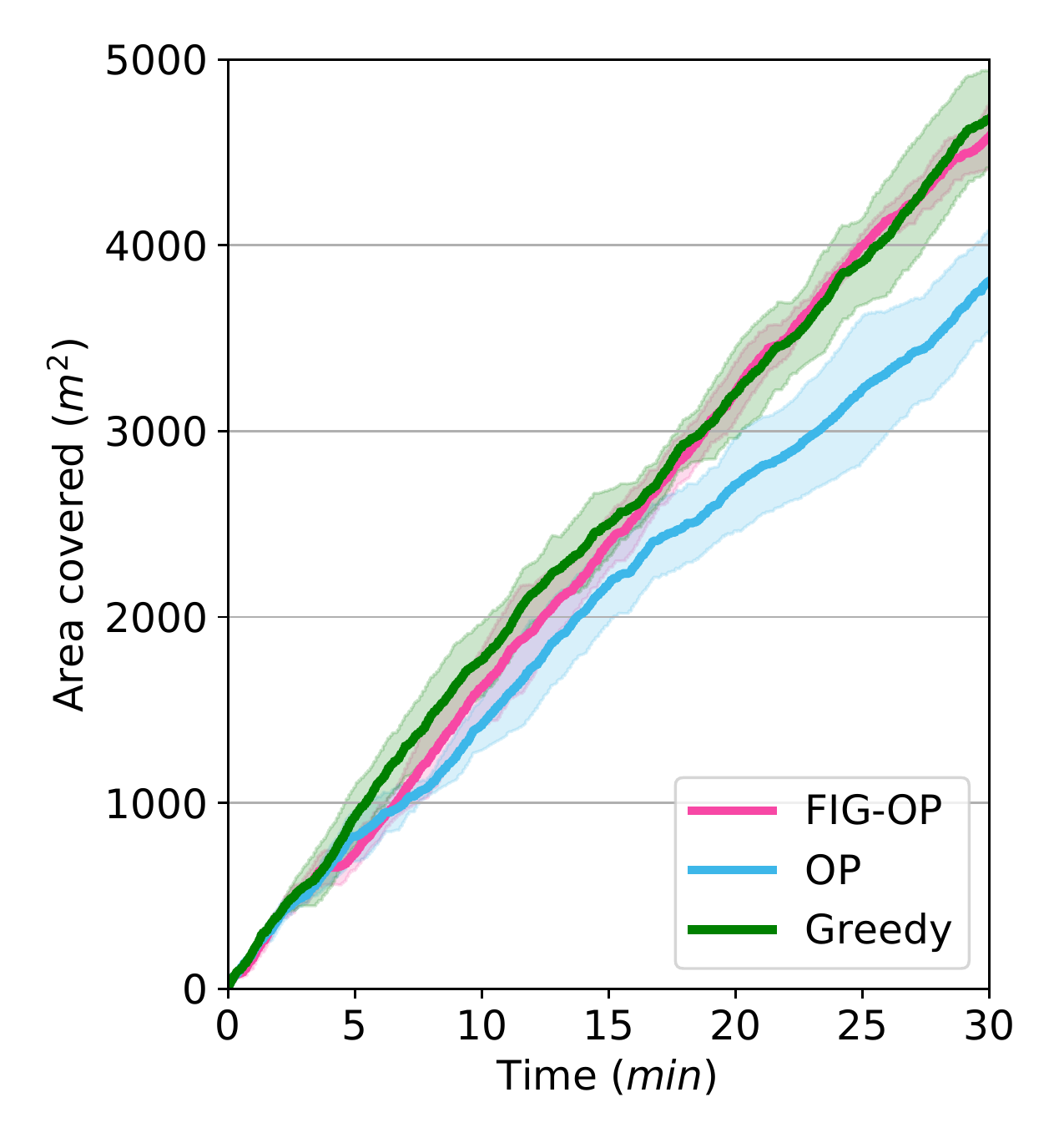}}	
    
	\caption{Exploration by our proposed \algname{} and baseline methods in a simulated maze. 
	The exploration metric for each method is averaged over five runs.}
    \label{fig:maze}
\end{figure}

\noindent Simulation results are provided in Figs. \ref{fig:subway}, \ref{fig:maze} and Table~\ref{table:maze_results}. 

\ph{Model Uncertainty Compensation}
The effectiveness of FIG-OP's greedy incentive is most evident in the maze environment. The maze consists of a large irregular network of passages. 
Most passages lead to long unexplored branches, a fact which is not encoded in the robot's model of the environment due to sensing limitations. As a result, in many cases the information gain assigned to a frontier underestimates the frontier's true value. 
Due to its false assumption of no model uncertainty, the OP method suffers in this setting since it plans over the full mission horizon by maximizing an unreliable reward estimate. 
On the other hand, both \algname{} and greedy approaches perform well since frontiers, leading to long unexplored branches, are visited earlier in the plan.

\ph{Long-Term Efficiency}
The effectiveness of FIG-OP's long-horizon planning is most evident in the subway environment. The subway consists of interconnected, polygonal rooms.
Near the beginning of the mission, the environment model is inaccurate since frontiers represent large swaths of space, and the model changes drastically as these frontiers are visited. Hence, FIG-OP and the greedy algorithm exhibit a higher coverage rate than OP, as shown in Fig.~\ref{fig:subway}c. As the mission progresses and frontiers no longer represent large areas (i.e. model uncertainty reduces), then the coverage rate for the greedy method decreases.
Meanwhile, OP and FIG-OP exhibit high coverage rates as they efficiency collect the remaining information in the environment.  
By strategically balancing greedy incentive and long-horizon efficiency, FIG-OP
explores 95\% of the subway faster (on average) than the greedy and OP methods, as shown in Fig.~\ref{fig:subway}b. 

\ph{Multi-Fidelity World Model}
We compare \algname{} with its low-fidelity action cost counterpart FIG-LF, Table~\ref{table:maze_results}. 
We find that integrating information from the metric map significantly improves coverage capabilities in both simulation environments, with a notable 35\% improvement in coverage rate in the maze environment.

\subsection{Field Tests}

\begin{figure}[t]

    \centering

    \subfloat[][]{%
      \includegraphics[clip,width=0.95\columnwidth]{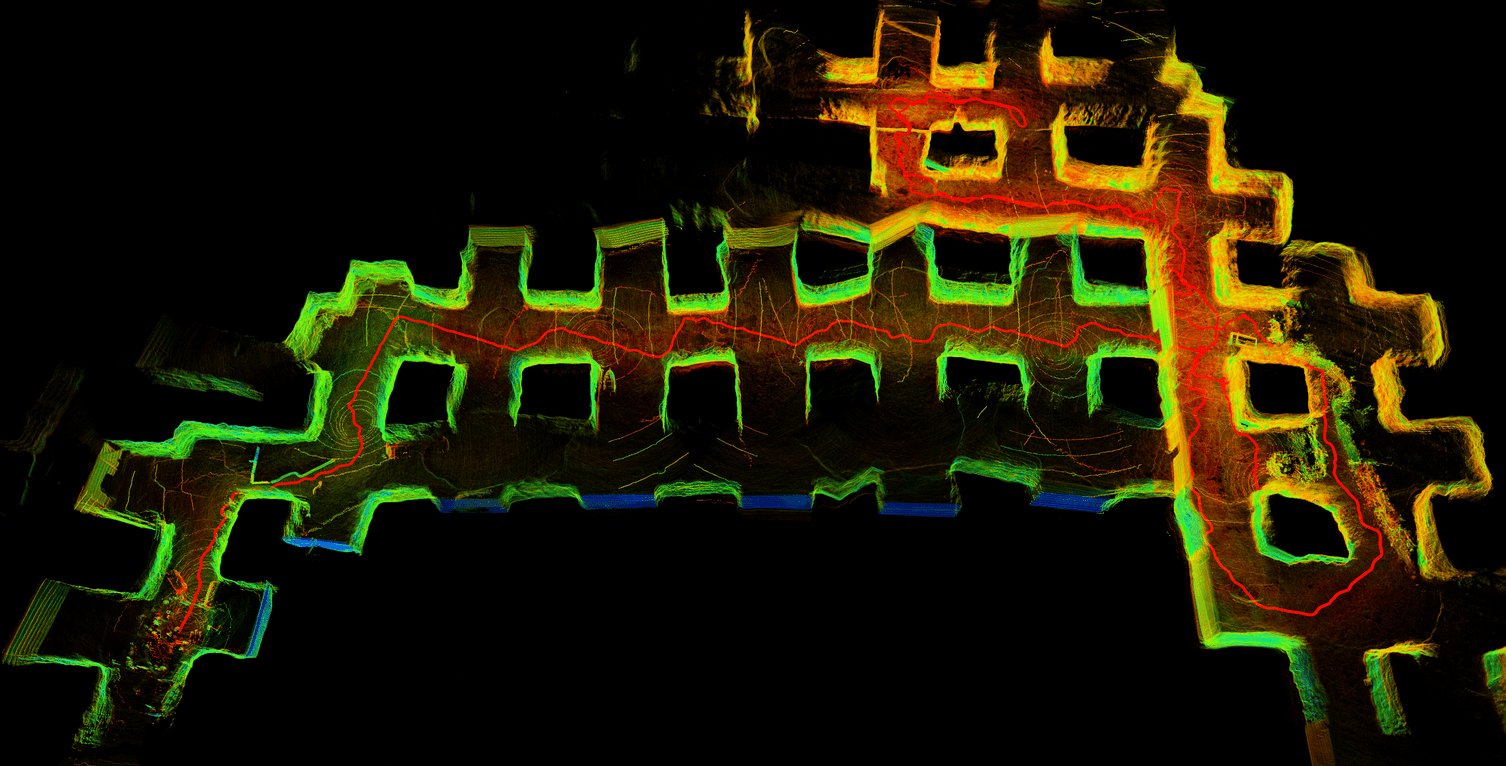}%
    } \\
    \subfloat[][]{%
      \includegraphics[clip,width=0.61\columnwidth]{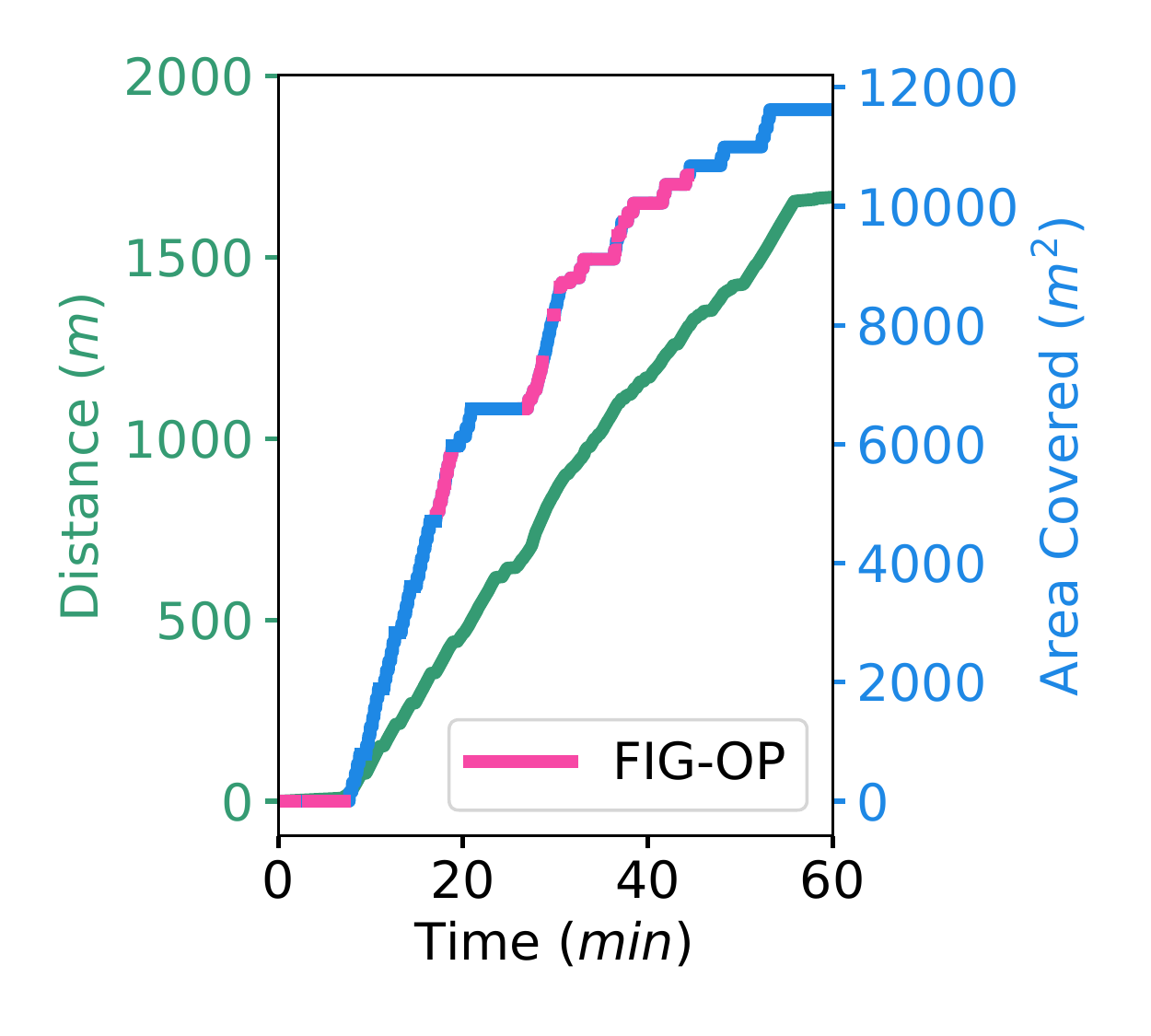}%
    }
    \subfloat[][]{%
    \includegraphics[clip,width=0.39\columnwidth]{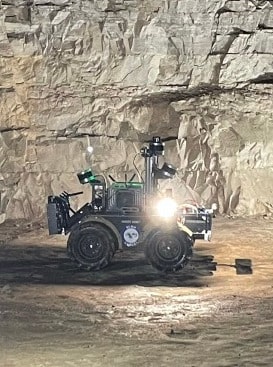}%
    } \,
    
    \caption{Autonomous exploration of a limestone mine by the Husky robot (c). The robot's trajectory is overlaid on the aggregated LiDAR point cloud (a). In the exploration metric (b), pink denotes time intervals where \algname{} was directly guiding the robot. Note that when the robot has been outside of the communications range of the base station for more than 10 minutes, the robot retraces its steps until communication is reestablished. 
    }
    \label{fig:hardware_husky_mine}
\end{figure}

We extensively tested our \algname{} solution on physical robots in a subway system, mine, and storage facility. During hardware tests, FIG-OP was integrated within the larger autonomy framework PLGRIM, introduced in \cite{sungpilgrim}. That is, the planning system onboard the robot alternates between a local viewpoint-based planner and a global frontier-based planner (i.e. \algname{}). 
Fig. \ref{fig:hardware_husky_mine} show the findings from an autonomous exploration run in a mine.

\ph{Solution Regularization}
The benefits of the solution regularization provided by FIG-OP is most notably demonstrated in a real-world environment where traversability risk estimates fluctuate. 
Using traversability risk estimates from the mine run (Fig. \ref{fig:hardware_husky_mine}), we evaluated path regularization based on different forms of the frontloading function $F$ in Eq.~(\ref{eq:FIG}). Our findings in Fig.~\ref{fig:sensitivity}b indicate that our proposed logistic form regularizes the solution over consecutive planning episodes. As a result, the robot can maintain continuous velocities, which is essential for rapid exploration. 

\section{Conclusion}\label{sec:conclusion}
We present a novel planning framework for autonomous exploration of large-scale complex environments under mission time constraints. Our proposed formulation FIG-OP is a generalization of the orienteering problem where the robot does not have access to a reliable environment model. 
FIG-OP compensates for model uncertainty by incorporating a greedy incentive that shifts information gain earlier in time. 
We solve FIG-OP over a multi-fidelity world representation, 
and demonstrate its ability to strike an effective balance between near- and long-term planning through an extensive test campaign. 

\section*{Acknowledgments}

We acknowledge our team members in Team CoSTAR for the DARPA Subterranean Challenge, and the resource staff at the Kentucky Underground facility.

\printbibliography

@String { icaps       = {International Conference on Automated Planning and Scheduling (ICAPS)} }

@String { icra        = {IEEE International Conference on Robotics and Automation (ICRA)} }

@String { iros        = {IEEE/RSJ International Conference on Intelligent Robots and Systems (IROS)} }

@String { rss         = {Robotics: Science and Systems} }

@article{choset2001coverage,
  title={Coverage for robotics---A survey of recent results},
  author={Choset, Howie},
  journal={Annals of Mathematics and Artificial Intelligence},
  volume={31},
  number={1},
  pages={113--126},
  year={2001},
  publisher={Springer}
}

@inproceedings{heng2015efficient,
  title={Efficient visual exploration and coverage with a micro aerial vehicle in unknown environments},
  author={Heng, Lionel and Gotovos, Alkis and Krause, Andreas and Pollefeys, Marc},
  booktitle=icra,
  pages={1071--1078},
  year={2015},
  organization={IEEE}
}

@article{singh2009efficient,
  title={Efficient informative sensing using multiple robots},
  author={Singh, Amarjeet and Krause, Andreas and Guestrin, Carlos and Kaiser, William J},
  journal={Journal of Artificial Intelligence Research},
  volume={34},
  pages={707--755},
  year={2009}
}

@inproceedings{faigl2013determination,
  title={On determination of goal candidates in frontier-based multi-robot exploration},
  author={Faigl, Jan and Kulich, Miroslav},
  booktitle={2013 European Conference on Mobile Robots},
  pages={210--215},
  year={2013},
  organization={IEEE}
}

@inproceedings{binney2010informative,
  title={Informative path planning for an autonomous underwater vehicle},
  author={Binney, Jonathan and Krause, Andreas and Sukhatme, Gaurav S},
  booktitle=icra,
  pages={4791--4796},
  year={2010},
  organization={IEEE}
}

@article{filliat2003map,
  title={Map-based navigation in mobile robots:: I. a review of localization strategies},
  author={Filliat, David and Meyer, Jean-Arcady},
  journal={Cognitive Systems Research},
  volume={4},
  number={4},
  pages={243--282},
  year={2003},
  publisher={Elsevier}
}

@article{thrun1998learning,
	Author = {Thrun, Sebastian},
	Journal = {Artificial Intelligence},
	Number = {1},
	Pages = {21--71},
	Publisher = {Elsevier},
	Title = {Learning metric-topological maps for indoor mobile robot navigation},
	Volume = {99},
	Year = {1998}}

@inproceedings{yamauchi1997frontier,
	Author = {Yamauchi, Brian},
	Booktitle = {IEEE International Symposium on Computational Intelligence in Robotics and Automation},
	Pages = {146--151},
	Title = {A frontier-based approach for autonomous exploration},
	Year = {1997}}

@inproceedings{keidar2012robot,
  title={Robot exploration with fast frontier detection: theory and experiments},
  author={Keidar, Matan and Kaminka, Gal A},
  booktitle={International Conference on Autonomous Agents and Multiagent Systems},
  pages={113--120},
  year={2012}
}

@inproceedings{umari2017autonomous,
  title={Autonomous robotic exploration based on multiple rapidly-exploring randomized trees},
  author={Umari, Hassan and Mukhopadhyay, Shayok},
  booktitle=iros,
  pages={1396--1402},
  year={2017},
}

@article{gonzalez2002navigation,
  title={Navigation strategies for exploring indoor environments},
  author={Gonz{\'a}lez-Banos, H{\'e}ctor H and Latombe, Jean-Claude},
  journal={International Journal of Robotics Research},
  volume={21},
  number={10-11},
  pages={829--848},
  year={2002},
  publisher={SAGE Publications Sage UK: London, England}
}

@article{Lauri2016planning,
  author  = {Mikko Lauri and Risto Ritala}, 
  title   = {Planning for robotic exploration based on forward simulation},
  journal = {Robotics and Autonomous Systems},
  year    = 2016,
  volume  = 83,
  pages   = {15--31}
}

@inproceedings{AutoSpot,
  title={{Autonomous Spot: Long-Range Autonomous Exploration of Extreme Environments with Legged Locomotion}},
  author={{Bouman$*$}, A and {Ginting$*$}, MF and {Alatur$*$}, N and Palieri, Matteo and Fan, David and Touma, Thomas and Pailevanian, Torkom and Kim, Sung and Otsu, Kyohei and Burdick, Joel and Agha-Mohammadi, Ali},
  booktitle=iros,
  year={2020},
}

@inproceedings{Ebadi2020,
author = {Ebadi, K and Chang, Y and Palieri, M and Stephens, A and Hatteland, A and Heiden, E and Thakur, A and Morrell, B and Carlone, L and Agha-mohammadi, A},
booktitle = icra,
title = {{LAMP}: Large-scale Autonomous Mapping and Positioning for Exploration of Perceptually-degraded Subterranean Environments},
year = {2020}
}

@inproceedings{Otsu2020,
author = {Otsu, Kyohei and Tepsuporn, Scott and Thakker, Rohan and Vaquero, Tiago Stegun and Edlund, Jeffrey A. and Walsh, William and Miles, Gregory and Heywood, Tristan and Wolf, Michael T. and Agha-mohammadi, Aliakbar},
booktitle = {IEEE Aerospace Conference},
title = {{Supervised autonomy for communication-degraded subterranean exploration by a robot team}},
year = {2020}
}

@inproceedings{fan2021step,
  title={{STEP}: {S}tochastic traversability evaluation and planning for safe off-road navigation},
  author={Fan, David D. and Otsu, Kyohei and Kubo, Yuki and Dixit, Anushri and Burdick, Joel and Agha-mohammadi, Ali-akbar},
  booktitle={arXiv preprint arXiv:2103.02828},
  year={2021}
}

@article{AliNeBula21,
  title={{NeBula}: Quest for Robotic Autonomy in Challenging Environments; {TEAM CoSTAR} at the {DARPA} Subterranean Challenge},
  author={Agha-mohammadi, Ali and {et al.}},
  journal={Journal of Field Robotics},
  year={2021},
}

@inproceedings{AliNeBula21arXiv,
  title={{NeBula}: Quest for Robotic Autonomy in Challenging Environments; {TEAM CoSTAR} at the {DARPA} Subterranean Challenge},
  author={Agha-mohammadi, Ali and {et al.}},
  booktitle={arXiv preprint arXiv:2103.11470},
  year={2021}
}

@inproceedings{sungpilgrim,
  title={{PLGRIM}: Hierarchical value learning for large-scale exploration in unknown environments},
  author={{Kim$*$}, Sung-Kyun and {Bouman$*$}, Amanda and Salhotra, Gautam and Fan, David D and Otsu, Kyohei and Burdick, Joel and Agha-mohammadi, Ali-akbar},
  booktitle=icaps,
  volume={31},
  pages={652--662},
  year={2021}
}

@article{howard2006experiments,
  title={Experiments with a large heterogeneous mobile robot team: Exploration, mapping, deployment and detection},
  author={Howard, Andrew and Parker, Lynne E and Sukhatme, Gaurav S},
  journal={The International Journal of Robotics Research},
  volume={25},
  number={5-6},
  pages={431--447},
  year={2006},
  publisher={SAGE Publications}
}

@inproceedings{tare,
  title={TARE: A Hierarchical Framework for Efficiently Exploring Complex 3D Environments},
  author={Cao, Chao and Zhu, Hongbiao and Choset, Howie and Zhang, Ji},
  booktitle=rss,
  year={2021}
}

@article{orienteering,
  title={The team orienteering problem},
  author={Chao, I-Ming and Golden, Bruce L and Wasil, Edward A},
  journal={European Journal of Operational Research},
  volume={88},
  number={3},
  pages={464--474},
  year={1996},
  publisher={Elsevier}
}

@article{gls,
  title={A guided local search metaheuristic for the team orienteering problem},
  author={Vansteenwegen, Pieter and Souffriau, Wouter and Berghe, Greet Vanden and Van Oudheusden, Dirk},
  journal={European Journal of Operational Research},
  volume={196},
  number={1},
  pages={118--127},
  year={2009},
  publisher={Elsevier}
}

@inproceedings{dang2019graph,
  title={Graph-based path planning for autonomous robotic exploration in subterranean environments},
  author={Dang, Tung and Mascarich, Frank and Khattak, Shehryar and Papachristos, Christos and Alexis, Kostas},
  booktitle=iros,
  pages={3105--3112},
  year={2019},
  organization={IEEE}
}

@article{fang2019autonomous,
  title={Autonomous robotic exploration based on frontier point optimization and multistep path planning},
  author={Fang, Baofu and Ding, Jianfeng and Wang, Zaijun},
  journal={IEEE Access},
  volume={7},
  pages={46104--46113},
  year={2019},
  publisher={IEEE}
}

@article{wang2019autonomous,
  title={Autonomous robotic exploration by incremental road map construction},
  author={Wang, Chaoqun and Chi, Wenzheng and Sun, Yuxiang and Meng, Max Q-H},
  journal={IEEE Transactions on Automation Science and Engineering},
  volume={16},
  number={4},
  pages={1720--1731},
  year={2019},
  publisher={IEEE}
}

@inproceedings{ester1996density,
  title={A density-based algorithm for discovering clusters in large spatial databases with noise.},
  author={Ester, Martin and Kriegel, Hans-Peter and Sander, J{\"o}rg and Xu, Xiaowei and others},
  booktitle={kdd},
  volume={96},
  number={34},
  pages={226--231},
  year={1996}
}

@inproceedings{burgard2000collaborative,
  title={Collaborative multi-robot exploration},
  author={Burgard, Wolfram and Moors, Mark and Fox, Dieter and Simmons, Reid and Thrun, Sebastian},
  booktitle={icra},
  volume={1},
  pages={476--481},
  year={2000},
  organization={IEEE}
}

@article{rouvcek2021system,
  author    = {Tom{\'{a}}s Roucek and
               Martin Pecka and
               {\v{C}}{\'i}{\v{z}}ek, Petr and
               Tom{\'{a}}s Petr{\'{\i}}cek and
               Jan Bayer and
               Vojtech Salansk{\'{y}} and
               Teymur Azayev and
               Daniel Hert and
               Matej Petrl{\'{\i}}k and
               Tom{\'{a}}s B{\'{a}}ca and
               Vojtech Spurn{\'{y}} and
               V{\'{\i}}t Kr{\'{a}}tk{\'{y}} and
               Pavel Petr{\'{a}}cek and
               Dominic Baril and
               Maxime Vaidis and
               Vladim{\'{\i}}r Kubelka and
               Fran{\c{c}}ois Pomerleau and
               Jan Faigl and
               Karel Zimmermann and
               Martin Saska and
               Tom{\'{a}}s Svoboda and
               Tom{\'{a}}s Krajn{\'{\i}}k},
  title     = {System for multi-robotic exploration of underground environments {CTU-CRAS-NORLAB}
               in the {DARPA} Subterranean Challenge},
  journal={arXiv preprint arXiv:2110.05911},
  year={2021}
}

@inproceedings{williams2020online,
  title={Online 3D Frontier-Based UGV and UAV Exploration Using Direct Point Cloud Visibility},
  author={Williams, Jason and Jiang, Shu and O’Brien, Matthew and Wagner, Glenn and Hernandez, Emili and Cox, Mark and Pitt, Alex and Arkin, Ron and Hudson, Nicolas},
  booktitle={IEEE International Conference on Multisensor Fusion and Integration for Intelligent Systems (MFI)},
  pages={263--270},
  year={2020},
  organization={IEEE}
}

@article{liu2021team,
  title={Team Orienteering Coverage Planning with Uncertain Reward},
  author={Liu, Bo and Xiao, Xuesu and Stone, Peter},
  journal={arXiv preprint arXiv:2105.03721},
  year={2021}
}

@article{yu2016correlated,
  title={Correlated orienteering problem and its application to persistent monitoring tasks},
  author={Yu, Jingjin and Schwager, Mac and Rus, Daniela},
  journal={IEEE Transactions on Robotics},
  volume={32},
  number={5},
  pages={1106--1118},
  year={2016},
  publisher={IEEE}
}

@inproceedings{visser2007beyond,
  title={Beyond frontier exploration},
  author={Visser, Arnoud and Ittersum, Merlijn van and Jaime, Gonz{\'a}lez and Luis, A and Stancu, Lauren{\c{t}}iu A and others},
  booktitle={Robot Soccer World Cup},
  pages={113--123},
  year={2007},
  organization={Springer}
}

\end{document}